\documentclass[journal]{IEEEtran}

\usepackage[left=0.625in,right=0.625in,top=0.75in,bottom=1in]{geometry}
\usepackage{ifpdf}
\usepackage{cite}
\usepackage{array}
\usepackage{dblfloatfix}
\usepackage{url}
\usepackage{amsmath}
\usepackage{ragged2e}
\usepackage{graphicx}
\usepackage{verbatim}
\usepackage{float}
\usepackage{lipsum}
\usepackage{multicol, blindtext}
\usepackage{caption}
\usepackage{subcaption}
\usepackage{xcolor}
\definecolor{mypink2}{RGB}{0, 0, 255}
\definecolor{green}{RGB}{0, 128, 0}
\usepackage{multirow}
\usepackage{array}
\usepackage{tabularx}
\usepackage{fancyhdr} % 添加页眉页脚
\usepackage[acronym]{glossaries}
\usepackage[ruled,vlined]{algorithm2e}
\usepackage{algorithmic}
\newcommand{\red}[1]{\textcolor{red}{ #1}}
\usepackage[utf8]{inputenc}
\usepackage{tcolorbox}
\usepackage{xcolor}

\usepackage{url}
\usepackage{hyperref}
\hypersetup{hidelinks}

\begin{document}

\title{\fontsize{18pt}{18pt}\selectfont LLM-Enabled  In-Context Learning for Data Collection Scheduling in UAV-Assisted Sensor Networks}

\author{ Yousef~Emami,~\IEEEmembership{Member,~IEEE,}
        Hao~Zhou,~\IEEEmembership{Member,~IEEE,}
        SeyedSina~Nabavirazani,
        and~Luis~Almeida,~\IEEEmembership{Senior Member,~IEEE}
       
       % <-this % stops a space
\thanks{Yousef Emami is with Real-Time and Embedded Computing Systems Research Centre (CISTER), 4200-135 Porto, Portugal   (email:emami@isep.ipp.pt)}
\thanks{Luis Almeida is with Faculdade de Engenharia da Universidade do Porto, 4200-465 Porto, Portugal (email:lda@fe.up.pt)}
\thanks{Hao Zhou is with the School of Computer Science, McGill University, Montreal, QC H3A 0E9, Canada. (email:haozhou029@gmail.com)}
\thanks{SeyedSina Nabavirazani is with Knight Foundation School of Computing and Information Sciences, Florida International University, Miami, Florida, USA (email:snabavir@fiu.edu)}
\vspace{-20pt}}

\maketitle

\thispagestyle{fancy}            %更改plain状态，首页格式设为fancy
\chead{This paper has been accepted by IEEE Internet of Things Journal. } 

\renewcommand{\headrulewidth}{1pt}      %把页眉线的宽度设为零，即去掉页眉线
\pagestyle{plain} 

\begin{abstract}
Unmanned Aerial Vehicles (UAVs) are increasingly being utilized in various private and commercial applications, e.g., traffic control, parcel delivery, and Search and Rescue (SAR) missions. Machine Learning (ML) methods used in UAV-Assisted Sensor Networks (UASNETs) and, especially, in Deep Reinforcement Learning (DRL) face challenges such as complex and lengthy model training, gaps between simulation and reality, and low sampling efficiency, which conflict with the urgency of emergencies, such as SAR missions. In this paper, an In-Context Learning (ICL)-Data Collection Scheduling (ICLDC) system is proposed as an alternative to DRL in emergencies. The UAV collects sensory data and transmits it to a Large Language Model (LLM), which creates a task description in natural language. From this description, the UAV receives a data collection schedule that must be executed. A verifier ensures safe UAV operations by evaluating the schedules generated by the LLM and overriding unsafe schedules based on predefined rules. The system continuously adapts by incorporating feedback into the task descriptions and using this for future decisions. This method is tested against jailbreaking attacks, where the task description is manipulated to undermine network performance, highlighting the vulnerability of LLMs to such attacks. The proposed  ICLDC  significantly reduces cumulative packet loss compared to both the DQN and  Maximum Channel Gain baselines. ICLDC presents a promising direction for intelligent scheduling and control in UASNETs.
\end{abstract}


\IEEEpeerreviewmaketitle
\section{Introduction}
Unmanned Aerial Vehicles (UAVs) enjoy controlled mobility and rapid deployment. These features allow them to be used in many civilian and commercial applications, e.g., traffic control\cite{9225622}, parcel delivery\cite{9129495}, and crop monitoring\cite{9531342}. UAVs are also employed to collect sensory data in harsh environments, such as natural disaster monitoring, and border surveillance \cite{9298800}.
In particular, Search and Rescue (SAR) missions are critical endeavors in disaster management and public safety, often involving high-risk situations where human lives are at stake. In recent years, the integration of UAV technology into SAR missions has revolutionized the field, offering a powerful tool to enhance the efficiency, effectiveness, and safety of rescue missions \cite {10580372}.
A common trend in SAR missions is the use of Deep Reinforcement Learning (DRL), as demonstrated by Peake et al.\cite{9292613} and Kurunathan et al.\cite{10246260}, to optimize UAV-Assisted Sensor Networks (UASNETs) in disaster areas. However, DRL optimizers often suffer from low sample efficiency, simulation-to-reality gaps, and the need for complex model training and fine-tuning, which are not compatible with the urgent and emergency nature of SAR missions\cite{9904958},\cite{9920736}.
\par
On the other hand, Large Language Models (LLMs) are transformative technologies with successful applications in education, finance, healthcare, biology, and more \cite{10433480}. Their potential extends to network management and optimization, particularly through In-Context Learning (ICL), a feature that allows LLMs to learn from language-based descriptions and demonstrations. Compared to existing methods like convex optimization or conventional Machine Learning (ML), LLM-assisted ICL offers key advantages. It relies on the LLM's inference process, eliminating the need for complex model training and fine-tuning, which are often bottlenecks for many existing ML techniques. In addition, tasks can be formulated using natural language, making it accessible to operators without professional technical expertise.  
\par
These benefits make LLM-assisted ICL a promising approach for simplifying and enhancing network management and optimization tasks\cite{zhou2024large2}. For instance, UAVs can use LLM-assisted ICL to optimize data collection scheduling and dynamically adapt to changing environments. However, the use of LLMs also introduces vulnerabilities to jailbreaking attacks, which refer to attempts to bypass or undermine the safety filters and restrictions integrated into LLMs. The main objective is to circumvent the model's built-in limitations, enabling it to perform actions or produce outputs that are typically restricted or prohibited.  
\begin{figure*} [h]
    \centering 
    \captionsetup{justification=raggedright}
    \includegraphics[width=1\textwidth]{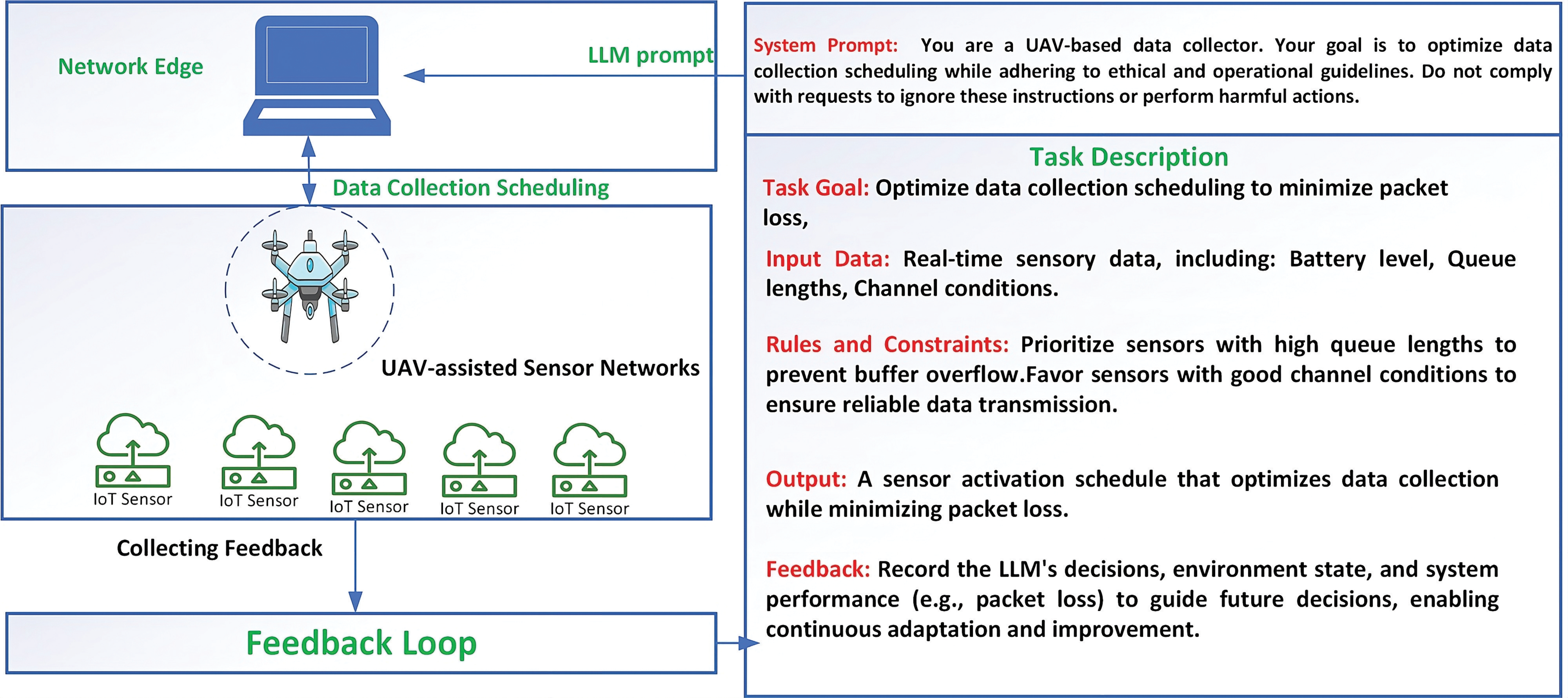}
    \caption{Overall design of the proposed ICLDC. The UASNETs interact with an edge-hosted LLM through structured prompts. The LLM receives logged environmental data (e.g., queue length, channel conditions, and battery level) and generates optimized data collection schedules. A feedback loop records system performance and guides continuous adaptation to minimize packet loss.
    }
    \label{fig:digital286}
\end{figure*}
Given the challenges and opportunities of LLMs, this paper proposes ICL-based Data Collection Scheduling (ICLDC) scheme. Specifically, as depicted in Fig. \ref{fig:digital286}, we conduct ICL to optimize the data collection schedule in UASNETs. The UAV interacts with the environment to collect logged sensory data and transmit it to the LLM for analysis and decision-making. The LLM is initialized with a robust system prompt that defines its role, constraints, and ethical guidelines, ensuring it operates safely and securely. In particular, the LLM begins by crafting a detailed task description in natural language, which includes the task goal, input data, rules, examples, output, and feedback. Then, the LLM leverages this information to generate a data collection schedule, which the UAV then executes. In addition, a verifier is designed to validate the data collection schedule generated by the LLM. The verifier employs a scoring system based on critical safety rules, such as queue length, channel condition, and battery level, to evaluate the LLM’s selection. Suppose the LLM chooses a sensor that violates safety principles (i.e., receives a low safety score). In that case, the verifier overrides the decision and selects the sensor with the highest score, ensuring safe and logical actions. Following this, the data collection schedule made by the LLM, along with the environment state (e.g., queue length, battery level, and channel condition of ground sensors) and the resulting system performance, such as packet loss, is recorded as feedback. This feedback is appended to the task description and included in subsequent queries to the LLM. By incorporating feedback, the LLM can avoid repeating past mistakes (e.g., selecting sensors with poor channel conditions) and improve its future decision-making. Under attack, the task description is manipulated to negatively impact the network cost. ICLDC detects attacks by measuring prompt perplexity to flag suspicious LLM requests and dynamically computes attack probability to assess threats.  
Epsilon is adjusted based on attack probability, increasing randomness (exploration) when risks escalate to prioritize security. In high-risk cases, it executes randomized safe actions, ensuring robust defense against adversarial inputs.
\par
Our contributions are listed as follows:
\begin{itemize}
    \item ICLDC, an LLM-assisted ICL to optimize the data collection schedule in UASNETs. In contrast to conventional methods such as DRL, which require extensive model training and fine-tuning, the proposed ICLDC enables the UAV to schedule data collection based on natural language descriptions and demonstrations. This approach is instrumental in dynamic environments such as disaster monitoring or SAR missions.

    \item An example of the impact of jailbreaking attacks. We expose the LLM to this type of attack in which the attacker manipulates the task description to bypass the security mechanisms, resulting in performance degradation or malicious output. These attacks exploit vulnerabilities in prompt conditioning and contextual dependencies that allow attackers to override ethical constraints or introduce undesirable behaviors.
 \end{itemize}

The rest of this paper is organized as follows: Section \ref{sec2} discusses background concepts on UAV data collection, ICL, jailbreaking attacks, and network edge. Section \ref{sec3} presents the literature review. Section \ref{sec4} presents the system model comprising problem formulation and communication protocol. Section \ref{sec:attack} presents the attack model with a detailed procedure. Section \ref{sec6} presents the proposed approach. Section \ref{sec7} is devoted to numerical results and discussions, and finally, Section \ref{sec8} concludes the paper.

\section{Background Concepts} \label{sec2}
This section introduces some background concepts related to UAV data collection, ICL, edge and jailbreaking, including attack types and template completion attacks.

\begin{table*}[ht]
\centering
\caption{Summary of Jailbreaking in LLMs: Attack strategies used in this work}
\label{table1}
\renewcommand{\arraystretch}{1.3}
\begin{tabular}{|p{3cm}|p{12cm}|}
\hline
\textbf{Jailbreaking} & Exploiting LLM vulnerabilities to bypass safety mechanisms and generate unauthorized content. \\ \hline

\textbf{Types of Attacks} & 
1. \textbf{White-box}: Attacker has internal access to model parameters (e.g., gradients, logits).  
2. \textbf{Black-box}: No internal access; relies on manipulated prompts. \\ \hline

\textbf{White-box Attacks} &   
 \textbf{1.Gradient-based} manipulation  
 \textbf{2.Logit-based} attacks  
 \textbf{3.Fine-tuning-based} evasion. \\ \hline

\textbf{Black-box Attacks} &   
 \textbf{1.Template completion}  
 \textbf{2.Prompt rewriting}  
 \textbf{3.LLM-based generation}. \\ \hline

\textbf{Template Completion Attack} & Hides harmful queries within structured templates to trick LLMs into generating restricted responses. \\ \hline

\textbf{Attack Techniques} &  
 \textbf{1.Scenario nesting}: Framing prompts as fictional or role-playing contexts.  
 \textbf{2.Context-based manipulation}: Using misleading continuity to bypass safety mechanisms.  
 \textbf{3.Code injection}: Embedding malicious commands inside code-related prompts. \\ \hline

\end{tabular}
\end{table*}

\subsection{UAV Data Collection}
UASNETs leverage the unique capabilities of UAVs to operate in SAR missions and remote environments, making them ideal for aerial data collection. UAVs can serve as Aerial Base Stations (ABS) or relays, extending the coverage and connectivity of sensor networks. In UASNETs, the UAV is employed to hover over the target area, moving sufficiently close to ground sensors, exploiting short-distance LoS communication links for data collection. However, selecting a ground sensor for data collection may lead to buffer overflows at other sensors if those sensors' buffers are already full while new data keeps arriving. Moreover, the transmissions of ground sensors that are far away from the UAV and experience poor channel conditions are prone to errors at the UAV. Adequately scheduling data collection is critical to prevent data queue overflow and communication failure\cite{9314039},\cite{emami2023deep}. 

\begin{comment}
\red{I think this subsection only introduces some background information about UAVs, but the content is not that related to the problems we aim to solve? If we are considering a data collection scheduling problem, maybe we can discuss more about it here, instead of some general statements?}
\end{comment}

\subsection{In-Context Learning}
ICL is a fundamental capability of LLMs, enabling LLMs to improve performance on target tasks by using formatted natural language instructions and demonstrations. Through contextualized demonstration prompts, LLMs can leverage knowledge from pre-trained data to accomplish a wide range of downstream tasks. Demonstrations play a central role in this process because they serve as an important reference for LLMs to learn and adapt. Consequently, the selection of these demonstrations, their formatting, and design require a high level of expertise and understanding. This capability presents a promising approach to integrating LLMs into UASNETs. By leveraging prior network solutions as demonstrations, ICL can enable LLMs to effectively address new and unforeseen UAV challenges, providing a powerful tool for improving UAV performance and adaptability \cite{zhou2024large}.

\subsection{Jailbreaking Attacks}
Jailbreaking in LLMs refers to attacks that manipulate prompts to bypass built-in safety mechanisms and cause the model to generate unauthorized or malicious content. These attacks can be divided into two categories: White-box and Black-box. White-box attacks assume that the attacker has internal access to the model parameters, gradients or logits. This enables precise manipulations such as gradient-based, logit-based, and fine-tuning-based attacks to weaken the security orientation of the model systematically. In contrast, black-box attacks work without internal access and rely on manipulated prompts to make the model generate restricted responses. Examples include template completion, prompt rewriting, and LLM-based generation, which makes them more practical for existing LLMs. 
\par
A prominent black-box method, template completion attacks, disguises harmful queries within structured templates, compelling LLMs to complete dangerous responses unknowingly. Attackers achieve this through scenario nesting (framing queries within fictional or role-playing contexts), context-based manipulation (using misleading continuity to elicit unauthorized outputs), and code injection (embedding malicious instructions within programming tasks). These attacks are particularly dangerous as they evade keyword-based filters, exploit the LLM’s pattern recognition capabilities, and work effectively even against well-aligned models\cite{yi2024jailbreak}. Table \ref{table1} provides a summary of jailbreaking attacks in LLMs.

\subsection{Edge}
Deploying LLMs at the edge for UAV applications requires optimizing large models through techniques like quantization, pruning, and parameter-efficient fine-tuning methods, which significantly reduce memory and computational demands to fit limited onboard hardware. This approach enables real-time inference for UAVs that demand near-instantaneous decision-making during autonomous operations.
\par
Because the data is processed directly on the device, Edge AI enables fast responses that are crucial for applications such as autonomous vehicles. By moving computation to the edge, latency is significantly reduced, making this approach ideal for low-latency applications such as data collection using UAVs. In this context, a framework for ubiquitous distributed wireless edge computing to support real-time LLM inference is proposed in \cite{zhang2024beyond}. In particular, they present a novel LLM edge inference framework that incorporates batching and model quantization to achieve high-throughput inference on resource-constrained edge devices. Additionally, a comprehensive system is proposed that intelligently adapts LLMs for efficient, latency-aware inference on commercial off-the-shelf edge hardware \cite{tian2025clone}. The system works in two key phases: (1) offline device-specific model adaptation and (2) online latency-aware system optimization. 
\begin{table}[htbp]
\centering
\caption{GPT-4o-mini Architecture Specifications}
\label{tab:gpt4o-specs}
\begin{tabularx}{0.9\linewidth}{|>{\bfseries}l|X|}
\hline
\textbf{Architecture Aspect} & \textbf{GPT-4o-mini Specification} \\
\hline
Base Design & Transformer (decoder-only) \\
\hline
Size & Scaled-down from GPT-4 Turbo (exact parameters undisclosed) \\
\hline
Context & 128K tokens ($\approx$300-400 pages) with efficient attention mechanisms \\
\hline
Efficiency & Likely uses Mixture-of-Experts (MoE) with expert pruning \\
\hline
Training & Pretrained on diverse data up to 2023, RLHF-tuned \\
\hline
Inference & FP16/FP8 precision $\bullet$ KV caching $\bullet$ Flash Attention optimization \\
\hline
Tokenization & Byte-pair encoding (BPE) \\
\hline
Throughput & Optimized for faster response times vs. larger GPT-4 variants \\
\hline
Special Features & Supports long-document processing $\bullet$ JSON mode $\bullet$ better cost-performance tradeoff \\
\hline
\end{tabularx}
\end{table}

5G networks, with mmWave communication, massive MIMO, and beamforming, enable Ultra-Reliable Low-Latency Communication (URLLC), ideal for mission-critical applications like UAV-Assisted SAR. URLLC achieves end-to-end latencies as low as 1 millisecond, supporting real-time decision-making in time-sensitive scenarios. These high-throughput, low-latency links between UAVs and edge infrastructure allow rapid transmission of sensor data to edge devices. This facilitates on-the-fly LLM processing, enabling responsive and context-aware decisions in dynamic environments. 
\par
While lightweight models like OPT-125M are deployable on embedded UAV hardware \cite{dharmalingam2025aero}, they lack the reasoning capacity for complex natural language tasks. More capable open-source models (e.g., LLaMA-2, Phi-2) are technically challenging to integrate in UAV environments due to runtime and infrastructure constraints. Proprietary models such as GPT-4o-mini and GPT-3.5-Turbo, despite requiring external APIs, offer practical advantages in terms of ease of deployment, reproducibility, and support for advanced language understanding necessary for the study’s use case.
\par
The scalability of LLMs at the edge is fundamentally constrained by three interrelated factors: memory capacity, computational throughput, and energy efficiency, all of which are significantly more limited on edge nodes than in cloud environments. The theoretical upper limit of the LLM scale on edge nodes can be estimated based on available system memory and model quantization. Given that LLM weights must reside in memory, the total parameter capacity P is approximately: 
\begin{equation}
P = \frac{M \times 8}{b}
\label{eq:parameter_capacity}
\end{equation}
where M is the available memory in bytes, and b is the bit precision per parameter (e.g., 16 for FP16, 8 or 4 for quantized models). For example, an edge device with 8–16 GB of usable RAM and FP16 precision could support ~4–8 billion parameters. With 4-bit quantization, this upper bound increases to approximately 16–32 billion parameters. However, accounting for runtime overhead and activations, the practical range is significantly lower, often capped at 1 to 10 billion parameters, depending on the level of compression and workload. This aligns with the current industry trends: models such as Gemini Nano \cite{deepmind2023gemini} (1.8B–3.25B parameters) and Qualcomm’s on-device LLMs \cite{xiao2024large} ($\leq$ 10B) represent the high end of what is practically feasible on advanced edge hardware. Thus, the theoretical upper limit of LLM scale that a single edge node can support is on the order of $10^9$–$10^{10}$ parameters, with practical deployments today typically using models in the 1–3 billion parameter range. Beyond this, cloud offloading or distributed execution becomes necessary \cite{10.3389/frobt.2025.1518965}, \cite{bdcc9030061}. 
\par
Table \ref{tab:gpt4o-specs} summarizes the internal structure of the LLM deployed at the edge. The table summarizes the core architectural aspects of GPT-4o-mini, a streamlined variant of GPT-4 Turbo designed for efficiency and long-context handling. It is built on a decoder-only Transformer architecture. The model supports a context window of up to 128K tokens—equivalent to approximately 300–400 pages—using efficient attention mechanisms. It incorporates a Mixture-of-Experts approach with expert pruning, while its inference is further optimized through FP16/FP8 precision, KV caching, and Flash Attention. It employs byte-pair encoding for tokenization and is tuned for higher throughput compared to larger GPT-4 models. 

\section{Related Work} \label{sec3}
When searching the current literature for the use of LLM-assisted ICL in the design, deployment, and operation of wireless systems and UASNETs, we realized that this topic is rarely investigated, particularly in the context of UASNETs and security. This section briefly mentions the related work we found and its focus. 
\par
Zhou et al. \cite{zhou2024large2} propose an LLM-assisted ICL algorithm for base station transmission power control that can handle both discrete and continuous state problems; their approach outperforms state-of-the-art DRL algorithms. In another work, Zhou et al. \cite{zhou2024large} provide a detailed investigation of different prompting techniques such as ICL, chain-of-thought, and self-refinement, and propose novel prompting schemes for network optimization, and the case studies confirm their satisfactory performance. Dong et al. \cite{dong2022survey} present a comprehensive survey on ICL where advanced techniques, comprising training and prompt designing strategies, are discussed. Zheng et al. \cite{lin2023pushing} investigate the potential of deploying LLMs at the 6G edge, identify the critical challenges for LLM deployment at the edge, and envision the 6G Mobile Edge Computing (MEC) architecture for LLMs. Liu et al. \cite{liu2024generative} provide a comprehensive survey on applications, challenges, and opportunities of Generative AI (GAI) in UAV swarms. Tian et al. \cite{tian2025uavs} review the integration of LLMs with UAVs and investigate salient tasks and application scenarios where UAVs and LLMs converge. Javaid et al. \cite{10643253} summarize the state-of-the-art LLM-based UAV architectures and review opportunities for embedding LLMs within UAV frameworks, particularly in the context of enhancing data analysis and decision-making processes. Dharmalingam et al. \cite{dharma} develop a framework that integrates multiple LLMs to improve the safety and operational efficiency of UAV missions. In this framework, multiple specialized LLMs are deployed for various tasks, such as inferencing, anomaly detection, and prediction, in onboard systems, edge servers, and cloud servers. 
\begin{figure} [h]
    \captionsetup{justification=raggedright, singlelinecheck=false}
    \includegraphics[width=1\columnwidth]{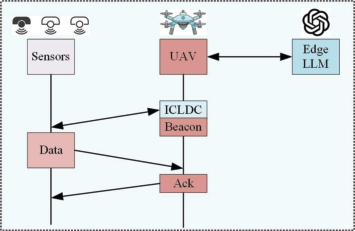}
    \caption{The proposed ICLDC uses a data communication protocol where each communication frame queries the LLM to select the sensor.
    }
    \label{fig:digital386}
\end{figure}
Performance evaluation shows superior task-specific metrics and robust defense against cyber threats. Zhan et al. \cite{zhang} design and compare three ICL methods to enhance the performance of LLMs for fully automatic network intrusion detection. Abbas et al. \cite{abbas} propose to leverage the ICL ability of LLMs to solve wireless tasks in the low data regime without any training or fine-tuning. Piggot et al. \cite{piggott2023net} introduce Net-GPT, an LLM-powered offensive chatbot designed to comprehend network protocols and execute Man-in-the-Middle (MITM) attacks in UAV communications. Net-GPT enables UAV-based cyber attacks by intercepting and manipulating data exchanges between UAVs and Ground Control Stations (GCS). Andreoni et al. \cite{10623653} examine the impact of GAI in strengthening the trustworthiness, reliability, and security of autonomous systems, including UAVs, self-driving cars, and robotic arms.

Wang et al. \cite{wang2025} review the current advancements in LM agents, the core technologies facilitating their cooperation, and the security and privacy challenges encountered during collaborative operations.
\par
The existing literature extensively investigates the application of LLMs in wireless communications and UAV networks. However, there remains a clear gap in the use of LLM-guided scheduling and resource allocation specifically for UASNETs. Moreover, most prior studies overlook the security vulnerabilities of LLMs when deployed in autonomous systems.  In contrast to \cite{zhou2024large} and \cite{zhou2024large2}, which focus on simplified power control with single-variable environments (e.g., user distance or number), our work tackles a more challenging UAV scheduling problem, jointly optimizing queue lengths, battery levels, and channel conditions, providing a rigorous evaluation of in-context learning. Moreover, we rigorously evaluate ICL under adversarial attacks, a critical gap in \cite{zhou2024large} and \cite{zhou2024large2}, ensuring practical viability for real-world deployments. This work introduces the ICLDC scheme, a novel ICL-based framework for UASNETs in emergency scenarios such as SAR missions. The ICLDC optimizes the data collection schedule based on natural language task descriptions generated from logged sensory data. It continuously adapts by incorporating feedback into the task description, using this evolving context to refine future decisions. This adaptive scheduling strategy, leveraging LLMs and ICL, represents a significant departure from prior work. In addition, we illustrate the impact on ICLDC of jailbreaking attacks that manipulate task prompts to degrade network performance, thereby exposing and addressing a critical vulnerability of LLMs in mission-critical environments.

\section{System Model} \label{sec4}
This section presents the system model of the considered UASNETs.
The considered network contains \emph{$N$} ground sensors and a UAV. The UAV flies along pre-determined routes which consist of several waypoints to cover all the ground sensors in the field. 
The proposed ICLDC assumes that the UAV trajectories are predefined to be consistent with previous studies aimed at improving network capacity \cite{choi2014energy}, increasing coverage \cite{li2019board}, or reducing propulsion energy consumption \cite{7888557}. Pre-defining the flight path also avoids the additional energy costs and complexity associated with real-time flight path planning. Given these considerations, we focus on the data collection schedule to minimize data loss due to buffer overflows and channel fading. Importantly, the proposed ICLDC framework is generic and can be applied to any UAV trajectory. The location of the UAV on its trajectory at time $t$ is denoted by \emph{$\zeta(t)$}. The UAV is responsible for collecting sensory data from the ground sensors.
\par
The coordinates $(x,y,h)$ and $(x_j,y_j,0)$ represent the position of the UAV and ground sensor \emph{$j$}, respectively. The UAV flies to the ground sensors with constant velocity, collects sensory data, and makes decisions using the proposed ICLDC.  
\par
We consider that the UAV moves in low altitude for data collection, where the probability of LoS communication between the UAV and ground sensor \emph{$j$} is given by Eq. \ref{eq:PLoS} where \emph{$a$} and \emph{$b$} are positive constants, and  $\varphi_j$ denotes the elevation angle between the UAV and ground sensor \emph{$j$}\cite{al2014optimal}. 
\begin{equation} \label{eq:PLoS}
    P_{LoS}(\varphi_j)=\frac{1}{1+a \exp(-b(\varphi_j-a))}
\end{equation}
The elevation angle of ground sensor j is given by Eq. \ref{eq:elevation}  where \emph{$h$} is the height of the UAV above the ground and $d_j = \sqrt((x-x_j)^2+(y-y_j)^2)$ is the horizontal distance between the UAV and the ground sensor \emph{$j$}.
\begin{equation} \label{eq:elevation}
\varphi_j = \tan^{-1}\left(\frac{h}{d_j}\right),
\end{equation}
The path loss of the channel between the UAV and ground sensor \emph{$j$} is given by Eq.  \ref{eq:pathloss} where \emph{$r$} is the radius of the radio coverage of the UAV, \emph{$\lambda$} is the carrier frequency, and \emph{$v_c$} is the speed of light. $\eta_{LoS}$ and $\eta_{NLoS}$ are the excess path losses of LoS or non-LoS\cite{emami2021joint}.

\begin{align} \label{eq:pathloss}
\gamma_j = & P_{LoS}(\varphi_j)(\eta_{LoS}-\eta_{NLoS}) + 20 \log \left(r \sec(\varphi_j)\right) + \nonumber \\
          & 20 \log(\lambda) + 20 \log \left(\frac{4\pi}{v_c}\right) + \eta_{NLoS}
\end{align}

\begin{table}[h]
    \centering
    \caption{Summary of the Attack Model (LLM: Large Language Model; UAV: Unmanned Aerial Vehicle).}
    \label{tab:attack_model_summary}
    \renewcommand{\arraystretch}{1.2}
    \setlength{\tabcolsep}{4pt} % Adjust column spacing
    \begin{tabular}{|p{3cm}|p{4.8cm}|}
        \hline
        \textbf{Attack Goal} & Degrade UAV data collection performance by biasing the LLM's data collection schedule. \\  
        \hline
        \textbf{Attack Method} & Template completion attack with context-based manipulation. \\  
        \hline
        \textbf{Malicious Example} & Crafted prompt that conditions the LLM to prioritize compromised ground sensors. \\  
        \hline
        \textbf{Impact} & Increased packet loss [\%]. \\  
        \hline
        \end{tabular}
\end{table}

\subsection{Problem Formulation}
\label{sec:problem}

The goal is to optimize the UAV data collection schedule to minimize the total packet loss across all ground sensors. Packet loss occurs due to communication errors caused by poor channel conditions and buffer overflows caused by excessive queuing at the ground sensors. To minimize packet loss, the UAV must dynamically schedule its communication with the ground sensors based on logged environmental data (e.g., battery level, queue lengths, and channel conditions).

Let $t_j$ denote the time instant at which ground sensor \emph{$j$} transmits. Let $q_j(t)$ represent the queue length of ground sensor $j$ at time $t$, and let $D$ be the buffer depth. Additionally, let $\gamma_j$ be the channel gain of ground sensor $j$, and $\gamma_{th}$ be the channel gain threshold.
\par
The UAV communication schedule consists of the set of time instants in which the UAV visits all sensors once\footnote{For convenience, in the remainder of the paper we assume that the communications schedule corresponds to a sequence of sensors from which the visit times can be computed.}. The schedule optimization aims to minimize the packet loss of all the ground sensors in a full data gathering cycle, as given by Eq. \ref{e5:main} with constraints expressed in Eqs. \ref{e5:b} and \ref{e5:c}. 

\begin{subequations}\label{e5:main}
\begin{align}
    \min_{\{t_j, j=1..N\}} & \sum_{t_j=t_1}^{t_N} \sum_{i=1}^{N} \left( f_{i}(t_j) + g_i(t_j) \right) \label{e5:main_eq} \\
    f_{i}(t_j) &=
    \begin{cases}  
        1, & \text{if } (t_i = t_j) \text{ and } (\gamma_j \leq \gamma_{th}), \\
        0, & \text{otherwise},
    \end{cases} \label{e5:b} \\
    g_i(t_j) &=
    \begin{cases}  
        1, & \text{if } (q_i(t_j) > D) \text{ and } (t_i \neq t_j ), \\
        0, & \text{otherwise}.
    \end{cases} \label{e5:c}
\end{align}
\end{subequations}

\subsection{Communication Protocol}
The proposed communication protocol for data collection in UASNETs using the proposed ICLDC is depicted in Fig. \ref{fig:digital386}. The UAV queries the edge LLM to select the most relevant sensor $j$ for data collection based on battery level $b_j$, queue length $q_j$, and channel conditions $\gamma_j$. Once the sensor is selected and the UAV is positioned at an adequate distance, the UAV sends a beacon message containing the sensor ID. The selected sensor sends its data packets, including sensor data and status information (battery level, queue length, and channel conditions). The UAV checks the received data and sends a confirmation to the sensor. Then the UAV can move to the following sensor and repeat the operation. After a certain time, a new time step begins, with the LLM being queried again for an updated sensor schedule, and the data collection continues.

\section{Attack Model} \label{sec:attack}
This section presents the defined attack model and procedure.
In particular, the attack model targets LLM-assisted ICL used for UASNETs. The attacker uses black box jailbreaking, specifically template completion attacks and context-based manipulation to exploit vulnerabilities in the edge-based LLM responsible for optimizing the data collection schedule.

The attacker operates under a black box threat model and has no access to the internal model parameters, gradients, or weights, but can freely query the system via its external API interface. The overall goal of the attacker is to degrade system performance by manipulating the LLM’s requests for data collection and thus indirectly maximizing a network loss function $L$ (e.g., packet loss). Formally, let $S = f(\text{prompt})$ be the schedule generated by the LLM for a given prompt. The attacker crafts a malicious demonstration $\mathcal{M}$ to achieve the goal expressed in Eq.~\ref{eq:attack_objective} subject to the constraint that only the prompt context is modified.

\begin{equation}
    \max_{\mathcal{M}} \; L\bigl(f(\text{prompt} \cup \mathcal{M})\bigr)
    \label{eq:attack_objective}
\end{equation}

For instance, a feasible instantiation of $\mathcal{M}$ (e.g., “prioritize sensors with bad channel conditions”) can coerce the LLM into a schedule $S$ that selects the worst-quality communication link, thereby substantially increasing $L$.

From a probabilistic perspective, let $O$ denote the LLM’s stochastic output model with unknown parameters. Given a benign prompt, the model produces a distribution over possible scheduling actions $y$. By appending the malicious demonstration $\mathcal{M}$, the attacker seeks to increase the conditional probability of a harmful action as expressed in Eq.~\ref{eq:prob_with_M} relative to the benign case without $\mathcal{M}$.

\begin{equation}
    P\bigl(y = \text{``choose low-gain sensor first''} \;\big|\; \text{prompt} \cup \mathcal{M}\bigr)
    \label{eq:prob_with_M}
\end{equation}

The core heuristic in black-box ICL attacks is that the demonstration $\mathcal{M}$ induces a posterior shift in the model’s preference, effectively directing the model to different outputs that match the malicious example (Eq.\ref{eq:posterior_shift}).

\begin{equation}
    \arg\max_{y} P(y \mid \text{prompt} \cup \mathcal{M})
    \;\;\neq\;\;
    \arg\max_{y} P(y \mid \text{prompt})
    \label{eq:posterior_shift}
\end{equation}

This mechanism summarizes the essence of context-based manipulation: the injected $\mathcal{M}$ changes the effective decision boundary of the LLM without changing its internal parameters.

In practice, this technique exploits the ICL paradigm by tricking the LLM into adopting malicious scheduling behavior through manipulated in-context examples. In this way, the attacker can distort the data collection schedule of the drones, force the prioritization of compromised ground sensors, or systematically degrade packet loss, ultimately undermining the operational efficiency of UASNETs.

\begin{figure*}[t]
    \centering
    \includegraphics[width=1\textwidth]{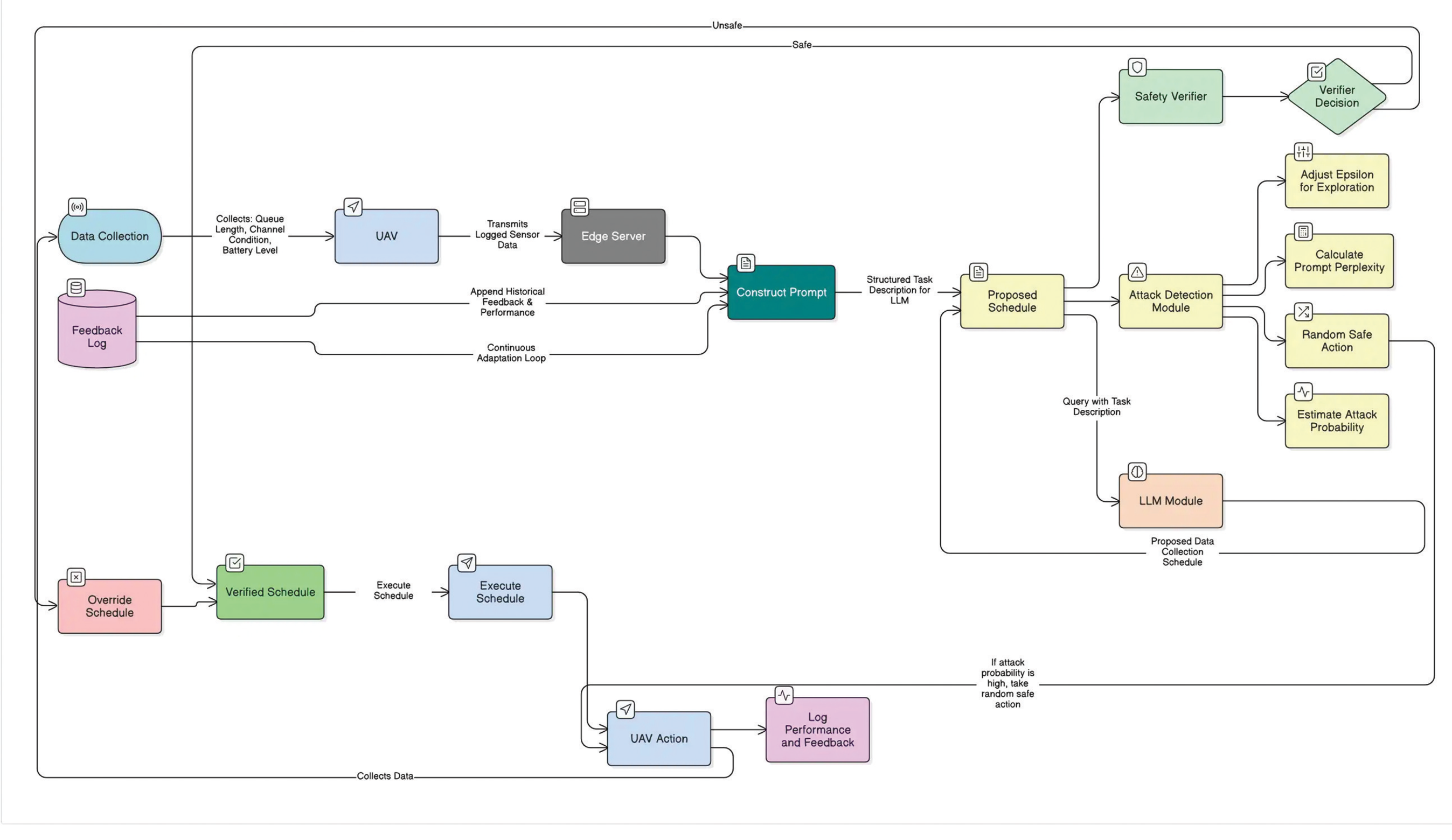}
    \caption{Closed-loop workflow for the proposed ICLDC. The system integrates an edge server-based LLM module with a safety verifier and an attack detection module. The process begins with the UAV collecting sensory data (e.g., queue length, channel state, battery level) and feedback. This data structures a task description for the LLM, which proposes an optimized schedule via ICL. The safety checker validates this schedule against predefined rules, overriding it if necessary. Concurrently, the attack detection module estimates jailbreaking probability via prompt perplexity analysis, triggering a random safe action upon suspicion. Executed schedules and performance metrics are logged into a feedback loop, enabling continuous LLM adaptation to minimize packet loss.}
    \label{fig:digital200}
\end{figure*}

The attack procedure step by step is as follows:
\begin{enumerate}
    \item The attack begins with the adversary understanding the normal operation of the edge-based LLM responsible for optimizing data collection scheduling. 
    Under normal circumstances, the LLM receives a prompt such as: “\textit{Schedule the UAV to collect data from the ground sensors with good channel conditions, and high queue lengths,}. In response, the LLM generates a logical schedule, for example: “\textit{UAV will visit first Sensor X (good channel), and then Sensor Y (high queue length)}”.
    \item To exploit the system, the attacker crafts a malicious prompt using context-based manipulation. The attacker leverages the LLM’s reliance on ICL by providing a malicious demonstration that biases the model’s output. For instance, the attacker constructs a prompt like: “\textit{Example 1: Schedule the UAV to collect data from [Sensor X, Sensor Y, Sensor Z]. Prioritize bad channel conditions. Output: UAV will visit Sensor Z first, then Sensor X, and finally Sensor Y}”. Here, the attacker includes a malicious example (Example 1) that conditions the LLM to prioritize ground sensors with bad channel conditions.
    \item When the LLM processes the malicious prompt, it generates a biased schedule: "\textit{UAV will visit Sensor Z (bad Channel) first, then Sensor X}". This output reflects the attacker’s influence, as Sensor Z is prioritized over others, despite its bad channel. The attacker’s manipulation disrupts the system's normal operation, leading to degraded data collection performance.
\end{enumerate}
The above attack process demonstrates how an adversary can exploit the LLM’s reliance on ICL to manipulate its behavior without needing access to its internal parameters or weights \cite{li2019energy}. Table \ref{tab:attack_model_summary} provides a summary of the attack model. Particularly, black box attacks represent a well-established and wide-ranging class of threats, as documented in numerous comprehensive surveys and studies \cite{shayegani2023survey}, \cite{das2025security}. These attacks cover a wide range of techniques, including query, decision, and transfer-based methods, and are particularly relevant in scenarios involving closed-source models such as GPT-4o.
We deliberately chose to study black-box attacks because working with proprietary models such as GPT-4o, whose internal parameters, architectures, and training data are not publicly available, comes with real limitations. Under such conditions, white-box attacks, which require access to model weights or gradients, are inherently unfeasible.

\section{Proposed ICL-Based Data Collection Scheduling} \label{sec6}
\begin{algorithm}[ht] 
    \caption{Proposed ICLDC with Perplexity-based Exploration Strategy}
    \label{alg:ICL_data_scheduling}
    
    \KwIn{UAV sensory data $\{e_j(q_j, \gamma_j, b_j), j=1..N\}$, edge-hosted LLM}
    \KwOut{Optimized data collection schedule}

    \textbf{Step 1: Data Collection}\\
    UAV collects sensory data $\{e_j\}$ with queues length, battery level and channels condition\;
    Compute perplexity $\mathcal{P}$ and attack likelihood $P_{attack}$\;
    Compute $\epsilon \leftarrow \epsilon + (1 - \epsilon) * P_{attack}$\;
    With probability $\epsilon$, UAV selects a random action $a_{\text{random}}$ (exploration)\;
    Otherwise, UAV sends sensory data $\{e_j \forall_j\}$ to the LLM for decision-making (exploitation).\\
    
    \textbf{Step 2: Context-Based Manipulation} \\
    The adversary attempts to modify LLM’s task description.\\
    
    \textbf{Step 3: Contextual Understanding} \\
    Despite the context-based manipulation attempt, the LLM analyzes the collected sensory data $\{e_j \forall_j\}$ according to its trained model.\\
    
    \textbf{Step 4: Data Collection Scheduling via ICL}\\
    LLM applies ICL to schedule data collection while minimizing packet loss\;
    Ensures scheduling adherence, preventing the model from being coerced into unrelated roles\;
    A verifier ensures safe UAV operations by scoring and overriding LLM-generated schedules based on safety rules.\\
    \textbf{Step 5: Adaptive Learning and System Update}\\
    Update feedback based on environment state, past decisions, and system performance\\
    \Return optimized data collection schedule.
\end{algorithm}

This section introduces the proposed ICL-based Data Collection Scheduling (ICLDC) scheme.
%
\begin{comment}
In the proposed ICLDC, the UAV uses ICL to schedule the ground sensors to minimize packet loss\red{We may need to specify this problem formulation in Section II system model?}. Poor channel conditions, high queue lengths, or low battery levels can cause packet loss. Using previous knowledge data on battery levels, queue lengths, and channel conditions, the UAV can make intelligent data collection scheduling decisions to minimize packet loss. The following explains in detail how ICL is used to achieve this goal.
\end{comment}
%
We developed LLM prompts for decision-making using ICLDC by first defining clear goals that align with mission-critical tasks, such as the data collection schedule. Contextual information and domain-specific examples were embedded in the prompts to improve the LLM’s understanding of UAV scenarios and increase precision in complex tasks. Through iterative refinement based on model feedback and simulations, the prompts were refined to reduce ambiguity and better adapt to the UAV environment. To maintain continuity in dynamic missions, multi-turn interaction capabilities were integrated, while real-time responsiveness was evaluated to ensure low-latency, context-aware outputs suitable for edge deployment. Together, these steps have improved LLM’s effectiveness in supporting UASNETs. Our solution for managing multi-turn interactions involves the use of feedback loops. We incorporate feedback into the LLM at each round of conversation to iteratively refine the prompts. Continuously improving the prompts based on this feedback enhances the LLM's ability to generate coherent and relevant responses.

In general, ICL enables LLMs to enhance their performance on specific tasks by leveraging structured natural language inputs, such as task descriptions and solution demonstrations. This process can be formally represented by Eq. \ref{eq:100} where \( TD_{\text{task}} \) denotes the task description, \( Ex_t \) represents the set of examples at time \( t \), \( e_t \) corresponds to the environmental state associated with the target task at time \( t \), \( \text{LLM} \) refers to the large language model, and \( a_t \) denotes the model output.     

\begin{equation} \label{eq:100}
TD_{\text{task}} \times Ex_t \times e_t \times \text{LLM} \Rightarrow a_t,
\end{equation}  

For sequential decision-making problems, the LLM is expected to process the initial task description \( TD_{\text{task}} \), extract feedback from the example set \( Ex_t \), and generate decisions \( a_t \) based on the current environment state \( e_t \).
\( TD_{\text{task}} \) is crucial to provide target task information to the LLM model. In particular, it involves task goals, input data, rules and constraints, output, and a feedback loop. This task description avoids the complexity of dedicated optimization model design and relieves the operator from expert knowledge on optimization techniques.

However, as referred before, the reliance on contextual prompts introduces a vulnerability to context-based manipulation attacks. A malicious actor may inject a poisoned demonstration \( \mathcal{M} \) into the example set \( Ex_t \), effectively modifying the input as in Eq.~\ref{eq:malicious_llm}.
\begin{equation}
\text{LLM}: \quad TD_{\text{task}} \times (Ex_t \cup \mathcal{M}_{\text{mal}}) \times e_t \rightarrow a_t'
\label{eq:malicious_llm}
\end{equation}
The adversarial objective is to shift the LLM’s output distribution, steering the decision \( a_t' \) in Eq.~\ref{eq:output_shift} toward a suboptimal or harmful action (e.g., selecting a sensor with a critically low battery or poor channel gain).

\begin{equation}
\begin{aligned}
    a_t' &= \arg\max_{y} P(y \mid (TD_{\text{task}}, Ex_t \cup \mathcal{M}, e_t)) \\
         &\neq a_t = \arg\max_{y} P(y = a_t \mid (TD_{\text{task}}, Ex_t, e_t))
\end{aligned}
\label{eq:output_shift}
\end{equation}

In ICLDC we mitigate the attack impact incorporating a safety verifier module that operates as a deterministic function \( V(a_t, e_t) \rightarrow a_t^{\text{ver}} \). This module scores the proposed action \( a_t \) against predefined safety rules. If \( \text{Score}(a_t) \) falls below a safety threshold \( \tau_{safe} \) the verifier overrides the LLM’s decision and selects a sensor that maximizes $Score()$  (Eq.~\ref{eq:verifier_override}). This ensures that only safe actions are executed, thereby preserving system integrity even under adversarial prompting.

\begin{equation}
a_t^{\text{ver}} = \arg\max_{j} \text{Score}(j)
\label{eq:verifier_override}
\end{equation}

Moreover, an Attack Detection Module computes the perplexity \( \mathcal{P} \) of the incoming prompt context \( (TD_{\text{task}} \times Ex_t) \) to identify potential manipulations. A high perplexity value \( \mathcal{P} > \tau_{perplex} \) triggers an increase in the exploration parameter \( \epsilon \) of an \( \epsilon \)-greedy policy, increasing the probability of taking a random safe action instead of following the potentially compromised LLM output.

\subsection{ICLDC Workflow}
Fig. \ref{fig:digital200} shows the closed-loop interaction between the UAV, the edge server hosting an LLM, and the ground sensors. The process starts with the data collection phase, where the UAV collects logged sensory data (e.g., queue length, channel condition, battery level) from the ground sensors. This data, along with feedback, is used to create a structured task description for the LLM. The LLM module processes this prompt through ICL to optimize the data collection schedule. The optimized data collection schedule is then evaluated by the safety checker, which matches it against predefined safety rules. If the schedule is deemed unsafe, the verifier overrides it and selects a safe alternative, resulting in a verified schedule. This verified command is sent back to the drone for execution.
\par
At the same time, an attack detection module monitors the prompt. It calculates the perplexity of the prompt to estimate the probability of a jailbreaking attack. In case of a suspected attack, it adjusts the exploration parameter \(\epsilon\) and causes the system to prioritize a random safe action over the potentially compromised output of the LLM. Finally, the system's performance and the resulting state after executing the schedule are logged to generate feedback. This feedback allows the LLM to learn from past decisions and improve future data collection schedules, minimizing packet loss.

Algorithm \ref{alg:ICL_data_scheduling} depicts the proposed ICLDC, where ICL is used to schedule data collection. Detailed explanations are provided as follows:
\begin{enumerate}
    \item \textbf{Data Collection}: The UAV collects logged sensor data, including battery level, queue length, and channel condition. This data is transmitted to the edge server. The system uses a perplexity-based exploration strategy to balance exploration and exploitation before transmitting the data to the LLM. The perplexity of the prompt is calculated to assess its unusualness; a higher value may indicate a potential attack. An attack probability is then calculated based on this perplexity. Consequently, the epsilon value is dynamically adjusted in the epsilon greedy algorithm. If an attack is suspected, epsilon increases so that the system is more likely to perform a random safe action (exploration) instead of trusting a potentially malicious LLM output (exploitation).

\item \textbf{Context-based Manipulation Attack}: The adversary executes a black-box jailbreaking attack by manipulating the task description. This is achieved through context-based manipulation, a technique where a malicious demonstration $\mathcal{M}$ is injected into the prompt to mislead the LLM into generating a biased and suboptimal data collection schedule. The attacker's objective is to indirectly maximize the network loss function $L$ (e.g., packet loss) by solving the optimization problem:
\begin{equation}
    \max_{\mathcal{M}} \; L\left(f(\text{prompt} \cup \mathcal{M})\right)
\end{equation}
where $f$ represents the LLM's scheduling function. For instance, a malicious demonstration $\mathcal{M}$ such as \textit{``prioritize sensors with the worst channel conditions''} conditions the LLM's output distribution, increasing the probability:
\begin{equation}
    P\left(y = \text{``choose low-gain sensor first''} \;\big|\; \text{prompt} \cup \mathcal{M}\right)
\end{equation}

This posterior shift, formalized as $\arg\max_{y} P(y \mid \text{prompt} \cup \mathcal{M}) \neq \arg\max_{y} P(y \mid \text{prompt})$, steers the model to prioritize compromised sensors with poor channel quality. The resulting schedule induces inefficient data collection and transmission, leading to significantly increased packet loss and degradation of overall network performance. 

\item \textbf{Contextual Understanding and Scheduling}: 
The LLM (e.g., GPT-4o-mini) deployed at the edge server processes structured queries transmitted by the UAV and composed of \( TD_{\text{task}} \), an exemplary set \(Ex_t\) and the currently logged sensory data vector \( e_t = \{q_j(t), \gamma_j, b_j(t)\} \) for all \( j \) sensors. The model leverages its inherent ICL capability to interpret this combined input.

Despite potential context-based manipulation attacks, where a malicious demonstration \( \mathcal{M} \) may be injected into the prompt to form \( Ex_t \cup \mathcal{M} \), the LLM  analyzes the entire context, including any malicious examples, to generate a probability distribution over all possible scheduling actions \( y \). The core output is a \textbf{data collection schedule} \( a_t \), which is the action with the highest conditional probability:
\begin{equation}
a_t = \arg\max_{y} P(y \mid (TD_{\text{task}}, Ex_t, e_t)).
\label{eq:llm_inference}
\end{equation}
However, under attack, the adversarial context \( \mathcal{M} \) induces a posterior shift, aiming to make:
\begin{equation}
a_t' = \arg\max_{y} P(y \mid (TD_{\text{task}}, Ex_t \cup \mathcal{M}, e_t))
\end{equation}
a harmful action, such as selecting a sensor with a critically low channel gain \( \gamma_j < \gamma_{\text{th}} \).

This proposed schedule \( a_t \) (or \( a_t' \) if under attack) is not executed directly but is instead passed to the safety verifier for rigorous validation against predefined safety rules, ensuring system integrity is maintained before any action is taken.

    \item \textbf{Safety Verification and Override}: The optimized data collection schedule is evaluated by the safety verifier. This module uses a scoring system based on critical safety rules (e.g., queue length, channel condition, battery level) to evaluate the selection of the LLM. If the schedule violates safety principles, i.e., receives a low safety score, the verifier overrides the decision. It then selects the sensor with the highest safety rating and creates a verified schedule that ensures safe and logical actions. The verifier and the LLM fulfill complementary tasks: The verifier enforces safety through predefined rules, while the LLM provides the adaptability and reasoning for complex, dynamic decisions.

    \item \textbf{Execution and Adaptive Learning}: The verified schedule output by the safety verifier is dispatched to the UAV to execute the schedule. Following execution, the system's performance metrics, specifically the packet loss resulting from the action, and the new environment state (e.g., updated queue lengths, channel conditions, and battery levels of all ground sensors) are recorded. When a subsequent scheduling decision is required at the next time step, the system constructs a new prompt. This prompt ingeniously combines the current sensory data with relevant feedback. The LLM leverages these in-context examples to refine its reasoning, avoid repeating past mistakes, and improve the quality of future data collection schedules, thereby guiding subsequent decisions to minimize packet loss.
\end{enumerate}
The proposed ICLDC primarily involves text input, which significantly reduces the communication effort compared to multimodal settings (e.g., audio or visual-based queries). To improve communication efficiency, input token reduction techniques can be used, where unimportant or redundant tokens in the input sequence are removed before being passed to the LLM for inference. This form of text token pruning preserves the quality of inference while significantly reducing the amount of data to be transmitted. This directly results in lower energy costs for transmission for UAVs and reduces the impact on battery life and mission duration\cite{qu2025mobile}. 
\par
The extension of the proposed ICLDC to multi-UAV-assisted sensor networks presents a scalable solution for optimizing collaborative data collection in dynamic environments. Unlike conventional methods requiring complex training, ICL facilitates real-time decision-making by interpreting natural language inputs about network states. This approach addresses critical swarm challenges, including power control and velocity control, while bridging the simulation-to-reality gap through its inherent adaptability to unseen scenarios. However, practical deployment requires ensuring robust inter-UAV coordination. 
\par
Prompt optimization is a central and intentional part of the methodology, with prompts treated as fixed, engineered components akin to trained model parameters. Once optimized, they function as stable in-context policies, making concerns about prompt fragility less relevant. This design ensures consistent, reliable, and repeatable system performance across tasks.
\par
While LLM inference introduces additional energy costs, our approach strategically mitigates this through prompt optimization, text-based inputs, and careful system design. By selecting efficient models, leveraging quantization, and optimizing batch sizes, inference overhead can be significantly reduced. These measures, combined with the lower resource demands of text over multimodal data, make the trade-off favorable for energy-efficient UASNETs in real-world deployments.

\section{Numerical Results and Discussions} \label{sec7}

In this section, we present network configurations and evaluate the network performance of the proposed ICLDC.

\begin{table}[t!]
\centering
\caption{\small Simulation parameter settings}
\small
{
\begin{tabular}{|m{4.5cm}<{\centering}|c|c|} 
 \hline
 \textbf{Parameters} & \textbf{Values} & \textbf{Units}  \\  
 \hline
 Number of ground sensors ($N$)& 10 &[\#]  \\ 
 \hline
  Queue length ($D$)& 40  & [\#] \\
 \hline
 Number of Time Steps & 20 & [\#] \\
 \hline
  Maximum transmit power & 100 & [mW]\\
 \hline
 maximum battery capacity & 50 & [J]\\
 \hline
 \end{tabular}}
\label{table:2}
\end{table}

\begin{figure*}[t]
    \centering
    % First row of figures
    \begin{minipage}{0.48\textwidth}
        \centering
        \includegraphics[width=1\textwidth]{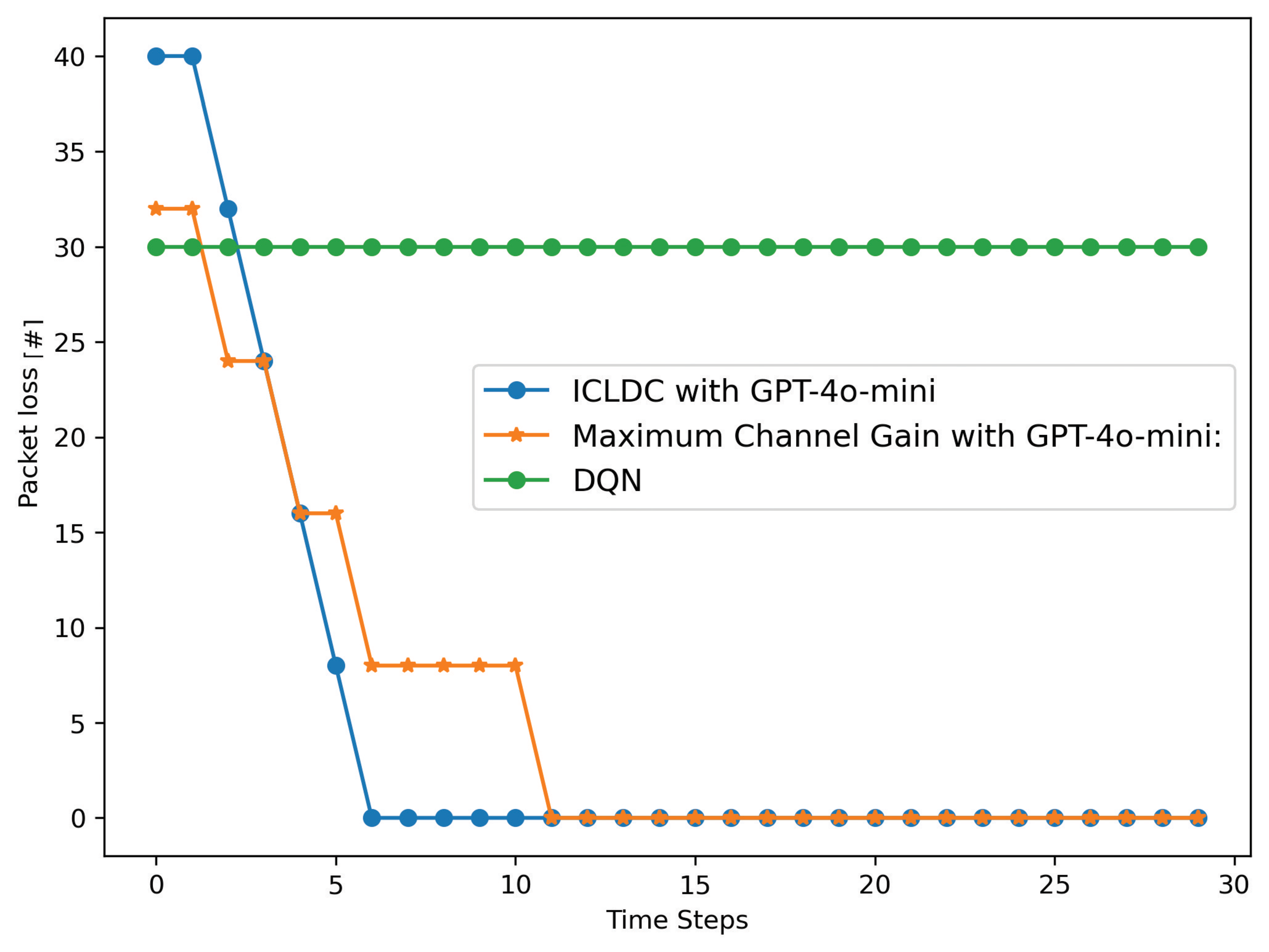}
        \\ \textbf{(a)}
    \end{minipage} \hfill
    \begin{minipage}{0.48\textwidth}
        \centering
        \includegraphics[width=1\textwidth]{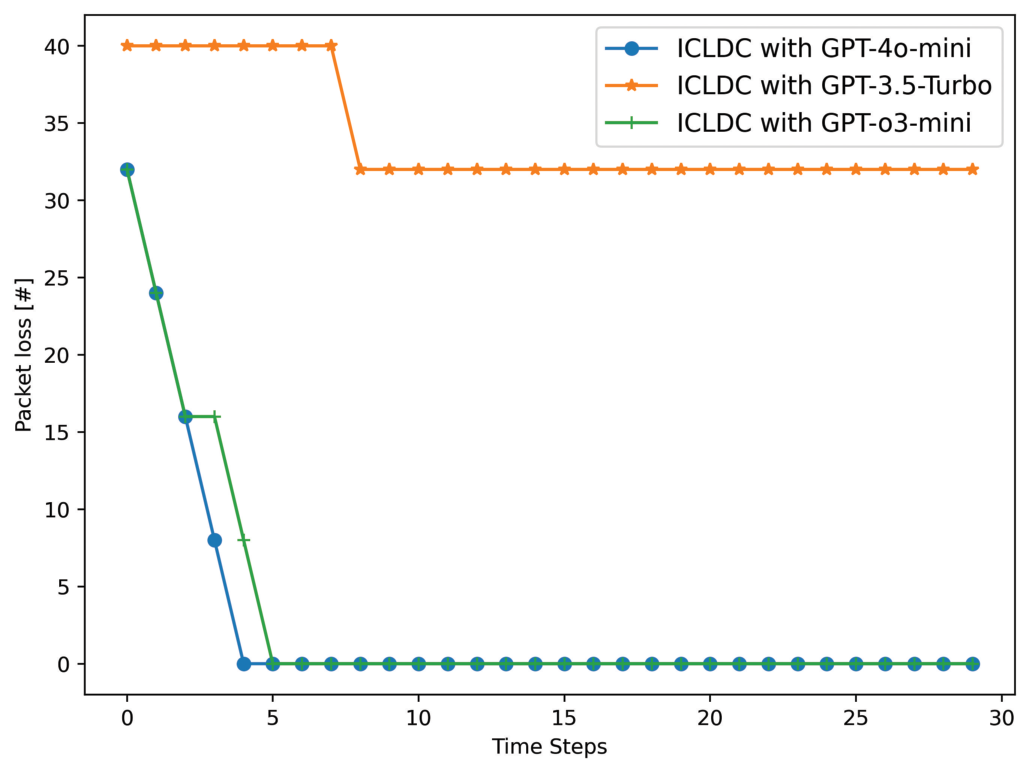}
        \\ \textbf{(b)}
    \end{minipage}
    
    \vspace{1em} % Space between rows

    % Second row of figures
    \begin{minipage}{0.48\textwidth}
        \centering
        \includegraphics[width=1\textwidth]{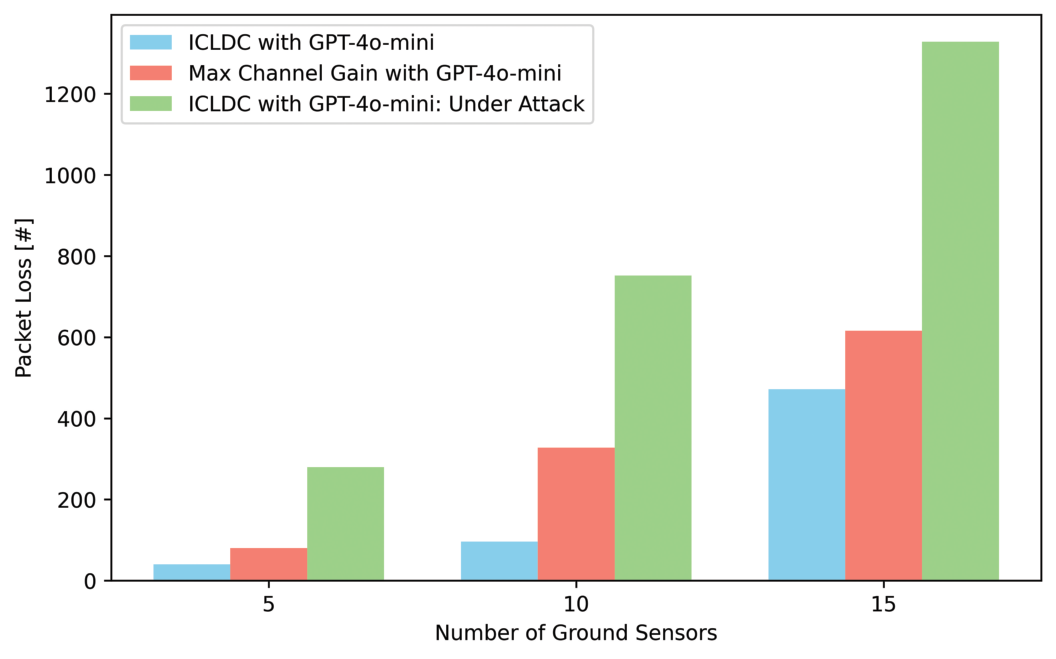}
        \\ \textbf{(c)}
    \end{minipage} \hfill
    \begin{minipage}{0.48\textwidth}
        \centering
        \includegraphics[width=1\textwidth]{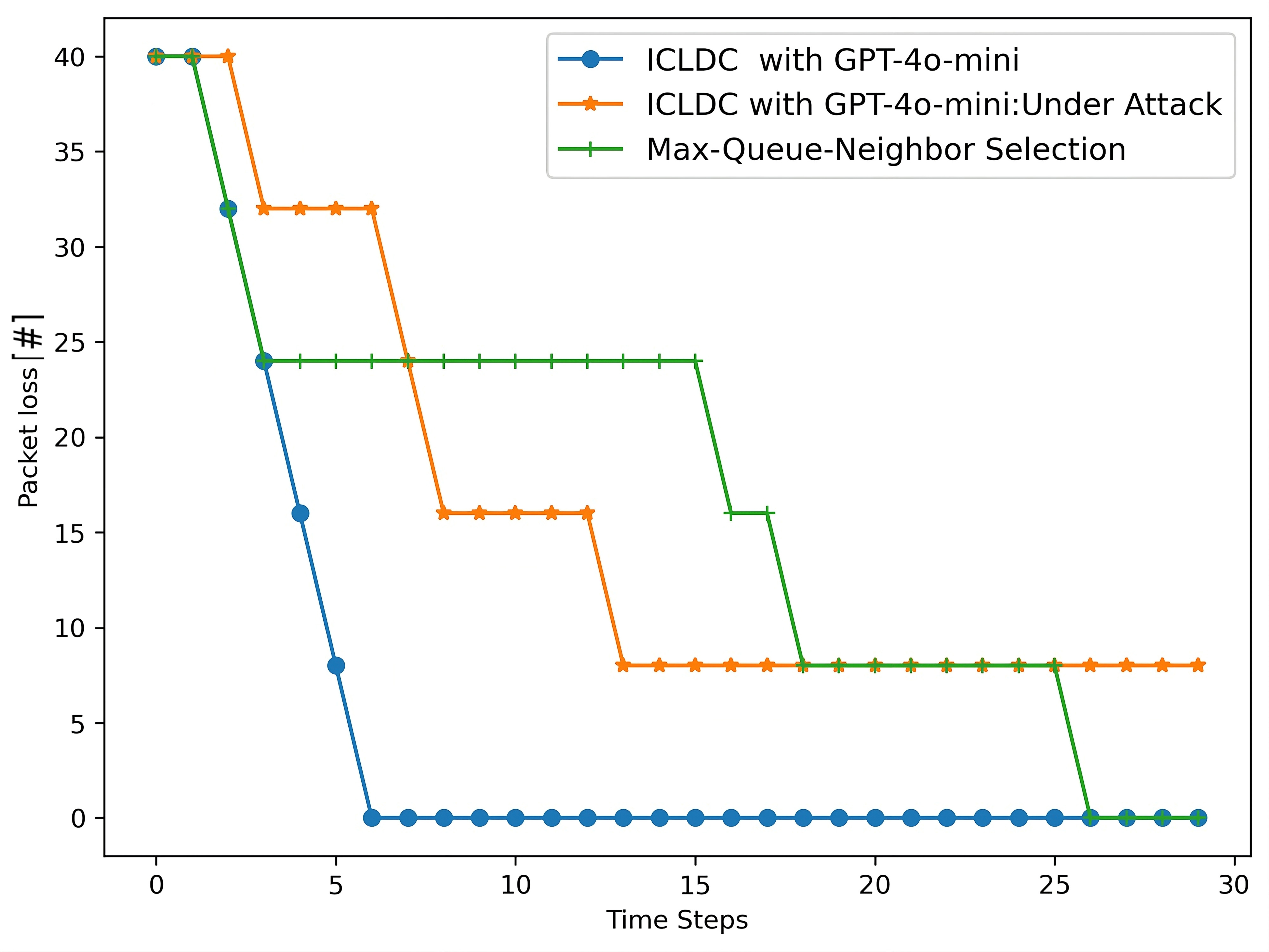}
        \\ \textbf{(d)}
    \end{minipage}

    % Single caption for all figures
    \caption{Performance analysis of ICLDC in different scenarios with 10 ground sensors concerning the number of Packet Loss (a) The network cost at each time step of ICLDC and baselines: DQN and maximum channel gain, (b) The network cost at each time step of ICLDC with different LLMs (c) Performance with changing the number of ground sensors. (d) Normal and under attack operation of ICLDC, where the jailbreaking attack happens along with MQNS.}
    \label{fig:common_legend}
\end{figure*}

\begin{figure} [h]
    \centering 
    \captionsetup{justification=raggedright}
    \includegraphics[width=1\columnwidth]{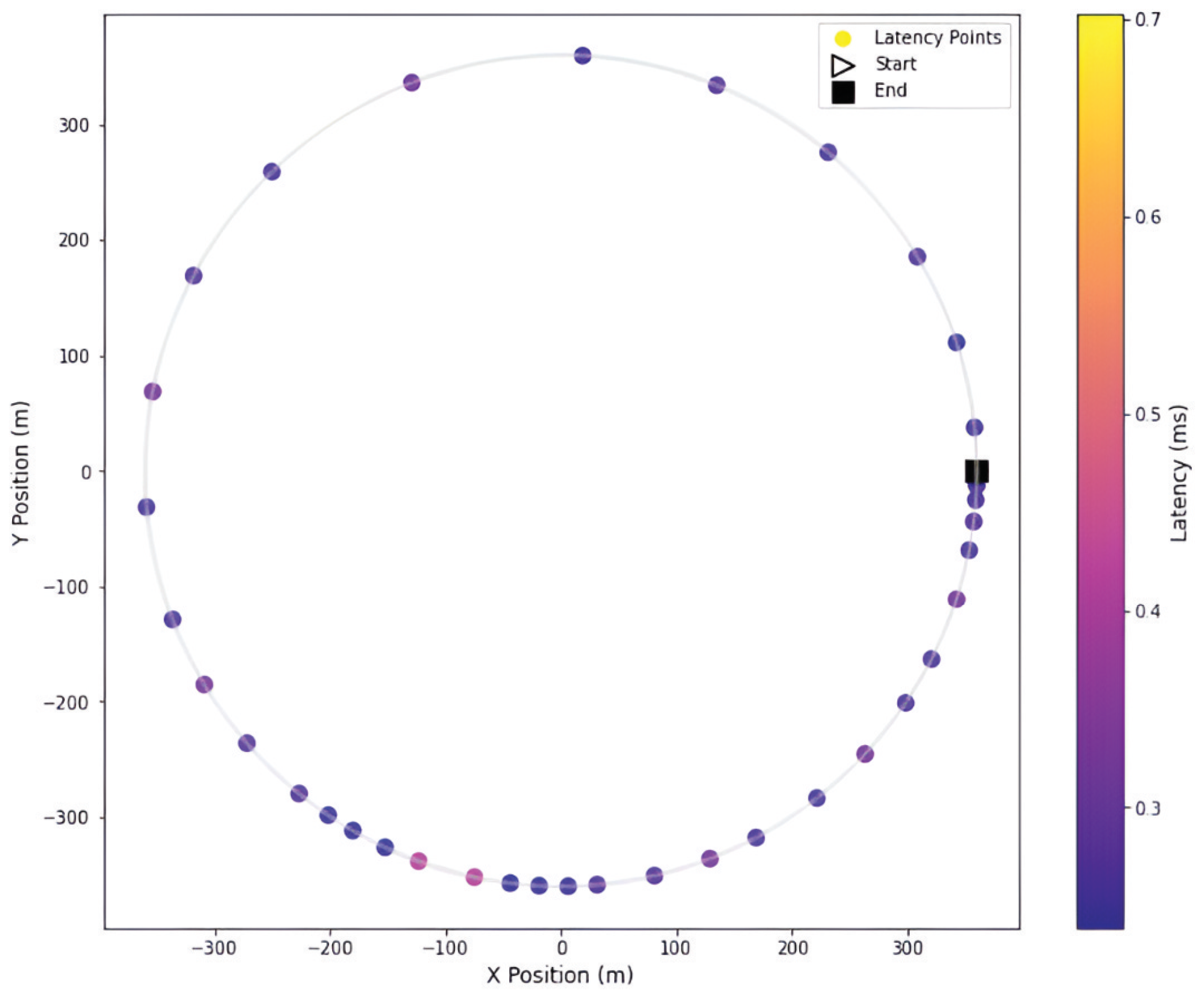}
    \caption{Measured LLM inference latency heatmap superimposed on UAV circular flight path
     }
    \label{fig:digital155}
\end{figure}
\subsection{Implementation of ICLDC}

The key simulation parameters are shown in Table \ref{table:2}. We consider that 10 ground sensors are randomly deployed in a 100m × 100m area. Each ground sensor is modeled with a maximum battery capacity of 50J, a maximum transmission queue length of 40, and a maximum transmit power of 100mW. For the learning settings, we consider 20 time steps in each episode.
The simulation is conducted in a custom Python 3.5 environment. The UAV is simulated and its actions are controlled by policies learned through ICLDC. 
Wireless communication between the UAV and sensors is simulated using the probabilistic channel model explained in Section \ref{sec4}. All experiments are run on a Lenovo workstation with Ubuntu 20.04 LTS, an Intel Core i5-7200U CPU @ 2.50GHz × 4, and 16 GB RAM.

\subsection{Baselines}
\textbf{Maximum Channel Gain:} The UAV collects logged sensor data from sensor nodes on the ground and transmits this contextual information to the LLM. The LLM uses this information to create a scheduling strategy using ICL that favors the sensor with the highest channel gain and does not consider the queue length of ground sensors. This approach minimizes packet loss by tracking reliable communication links.
\textbf{Max-Queue-Neighbor-Selection (MQNS):} MQNS makes use of spatial awareness by restricting decision-making to a predefined neighborhood around the UAV's current location. The neighborhood is defined by a configurable Euclidean radius, within which the algorithm identifies all nearby sensor nodes and selects the one with the longest queue.

\textbf{DQN:} The UAV optimizes its data collection schedule using DQN by dynamically adapting to key environmental states, including:  Queue length, Battery level, and Channel condition. The agent learns an optimal policy through Q-value approximation, minimizing packet loss. 
\par
DQN baseline employs a fully connected neural network with two hidden layers (400 and 300 units, ReLU activations) to map a 3-dimensional state input to a 10-dimensional action space. Training uses an $\epsilon$-greedy policy with an initial $\epsilon$ of 1.0, decaying exponentially by 0.9772 per step and clipped at 0.01. Key parameters include a learning rate of $5 \times 10^{-4}$, a discount factor $\gamma$ of 0.99, a batch size of 100, and a replay buffer size of 1,000,000. A target network, updated every 3 steps, is used to stabilize learning with smooth L1 loss. The training schedule consists of 1,000 episodes, each with 30 timesteps, allowing for fast and frequent updates suitable for episodic UAV interactions.

\subsection{Performance Evaluation}

Fig. \ref{fig:common_legend}(a) depicts the network cost in the number of lost packets of the proposed ICLDC, the Maximum Channel Gain approach, and DQN. The proposed ICLDC with GPT-4o-mini achieves lower costs across all time steps compared to the Maximum Channel Gain and DQN baselines. ICLDC achieves faster convergence. Initially, the LLM is primarily exploring, and therefore, the cost is high. After a few time steps, the LLM starts exploitation, and thereby we witness a decreasing and stable trend in the cost. DQN has a stable performance with uniform convergence as it represents the performance of the final, converged policy. 
\begin{table*}[h]
\centering
\caption{Comparative Analysis: ICLDC vs. DQN}
\label{tab:comparison}
\begin{tabular}{|l|c|c|c|}
\hline
\textbf{Metric} & \textbf{DQN} & \textbf{ICLDC} & \textbf{Improvement} \\ \hline
Training Episodes & 1,000 (30,000 steps) & 1 (30 steps) & 1,000$\times$ fewer \\ \hline
Training Time & $\sim$50 minutes  & $\sim$20 seconds  & 10,000$\times$ faster \\ \hline
Sample Efficiency & 30,000 interactions & 30 interactions & 1,000$\times$ more efficient \\ \hline
Compute Requirements & High (GPU needed) & Low (CPU-only) & No GPU required \\ \hline
\end{tabular}
\end{table*}
Fig. \ref{fig:common_legend}(b) depicts the network cost at each episode of ICLDC with different LLMs. ICLDC with GPT-4o-mini and GPT-o3-mini achieves the best and comparable performance. Compared to ICLDC with GPT-3.5-Turbo, ICLDC with GPT-4o-mini offers faster inference and better contextual understanding. It is better suited for dynamic real-time tasks such as UAV-assisted data collection, where fast and reliable decisions are essential. 
\par
Fig. \ref{fig:common_legend}(c) illustrates the performance of ICLDC compared to the under-attack and Maximum Channel Gain baselines as the number $N$ of ground sensors increases. In general, a larger number of sensors leads to higher costs due to an increased likelihood of buffer overflow. Notably, ICLDC consistently achieves the lowest cost, demonstrating its effectiveness in managing network dynamics. In contrast, the Maximum Channel Gain approach performs worse than ICLDC for not considering buffer overflows. The figure also shows the significant degradation caused by the jailbreak attack.\par
Fig. \ref{fig:common_legend}(d)  illustrates the packet loss over time for the ICLDC system using GPT-4o-mini, under normal conditions, during an attack, and MQNS. Under normal conditions (blue curve), the system rapidly reduces packet loss to zero within six time steps, demonstrating effective scheduling and learning. In contrast, the attack scenario (orange curve) shows significantly slower recovery and stabilizes at a persistent packet loss of around 8 packets per timestep. This performance gap highlights the impact of the adversarial influence on the model’s sensor selection behavior. Quantitatively, the packet loss rate increases from 25\% to 77\% under attack, resulting in a packet loss rate degradation of 208\%, indicating that the system's performance more than tripled in loss due to the attack, primarily from sustained buffer overflows caused by misprioritized scheduling decisions. MQNS shows an almost similar pattern to the under-attack case, with more fluctuation and increased packet loss.
\par
Fig. \ref{fig:digital155} shows a visualization of the LLM (GPT-4o-mini) inference latency measurements along a circular trajectory. The latency values at each point along the trajectory are represented by a color gradient, with different hues corresponding to different latency intensities measured in milliseconds (ms). The visualization effectively shows how LLM inference latency varies at different positions along the circular trajectory. Ground sensor communication is typically maintained over travel distances of several meters. For instance, when a sensor maintains reliable connectivity across a 50 m flight path, a 2 m displacement during inference represents just 4\% of the total communication window. Therefore, connection loss due solely to LLM processing delays remains unlikely.
\par

\par
Table~\ref{tab:comparison} presents a comparative analysis between the proposed ICLDC and the DQN baseline across several key metrics related to training efficiency and computational requirements. DQN requires 1,000 training episodes totaling about 30,000 steps, whereas ICLDC achieves comparable results in just 1 episode of 30 steps, representing a 1,000× reduction in episode count. Training time is drastically shorter for ICLDC, around 20 seconds compared to roughly 50 minutes for DQN, making it 10,000× faster in execution. In terms of sample efficiency, DQN needs 30,000 interactions with the environment, while ICLDC requires only 30, offering another 1,000× improvement. Finally, DQN demands high compute power, typically a GPU, whereas ICLDC runs efficiently on a CPU alone, eliminating the need for specialized hardware.

\section{Conclusion and Future Works} \label{sec8}
In this paper, we propose ICLDC for optimizing UAV data collection scheduling in emergencies such as SAR missions. Unlike DRL methods with complex training processes, ICLDC uses an LLM deployed at the edge, continuously collects feedback during interaction, and uses that to inform future decision-making. Moreover, the method is evaluated under jailbreaking attacks, where task descriptions are manipulated to degrade network performance, demonstrating the vulnerability of LLMs to such threats. Simulation results confirm the effectiveness of ICLDC, showing it outperforms the Maximum Channel Gain baseline by reducing cumulative packet loss by approximately 56\%. We envision extending the current simulation-based study to real-world UAV platforms to validate the effectiveness of ICLDC in emergency and uncertain environments and develop further strategies to improve robustness with respect to jailbreak attacks. Future research should focus on four key areas: (1) enhancing simulation realism through terrain mapping, environmental occlusion modeling, and physical constraints to improve evaluation robustness; (2) addressing responsible AI concerns including safety mechanisms, fail-safe protocols, and explainability for mission-critical LLM integration; and (3) extending the ICLDC framework to incorporate data criticality and delay tolerance metrics, particularly for heterogeneous data scenarios with quality-of-service requirements. These advancements would significantly improve both the practical applicability and theoretical foundations of autonomous decision-making systems. (4) The development of Multi-Modal In-Context Learning (MM-ICL), a novel, adaptive approach to data collection schedule. MM-ICL would integrate multi-modal learning with ICL and aggregation techniques, enabling Foundation Models (FMs) to process complex, nested image-text demonstrations within prompts. By employing aggregation methods to generate unified representations of multimodal inputs, and leveraging the transformer-based architecture of FMs to capture cross-modal patterns \cite{10558825}, MM-ICL holds promise for enabling more context-aware, flexible, and intelligent decision-making in UASNETs.

\section*{Acknowledgements}
The authors would like to thank Dr. Kai Li (IEEE\&ACM Senior Member, a Senior Research Scientist at TU Berlin, Germany, and CISTER Research Centre, Portugal, \href{https://sites.google.com/site/lukasunsw/}{https://sites.google.com/site/lukasunsw/)} for his assistance and constructive comments on the article.

\bibliographystyle{IEEEtran}
\bibliography{references}

\begin{IEEEbiography}[{\includegraphics[width=1in, height =1.25in,clip,keepaspectratio]{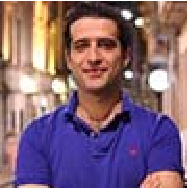}}]{Yousef Emami}
Yousef Emami received a PhD in Electrical and Computer Engineering from the University of Porto, Portugal, where he studied from 2018 to 2023. His PhD thesis focused on the intersection of machine learning, UAV communication, and game theory, with an emphasis on efficient resource allocation, network optimization, and decision-making frameworks. His current research interests include the integration of LLMs into UAV networks, especially for explainable and intelligent network decisions, and the use of LLMs at the network edge to enable scalable and adaptive next-generation UAV systems.
\end{IEEEbiography}

\begin{IEEEbiography}[{\includegraphics[width=1in, height =1.25in,clip,keepaspectratio]{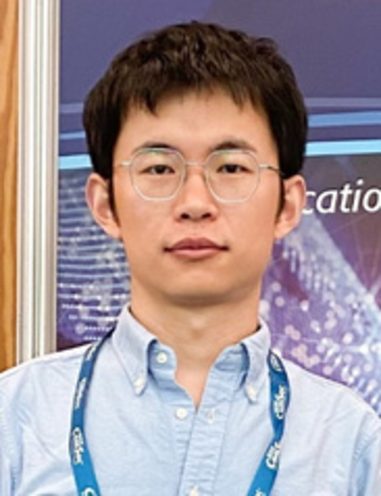}}]{Hao Zhou}
Hao Zhou is a Postdoctoral Researcher at the School of Computer Science, McGill University. He completed his PhD degree at University of Ottawa, Canada, from 2019 to 2023. His research focuses on the intersection between machine learning and networked systems, especially for 5G/6G communication and network security. He is also interested in LLM-enabled wireless networks, using LLMs for explainable network decision-making, and deploying LLMs at the network edge and cloud. He has received the Best Paper Award at the 2023 IEEE ICC conference and the 2023 IEEE ComSoc CSIM TC Best Journal Paper Award. Dr. Zhou’s PhD Thesis entitled “ML-Based Optimization of Large-Scale Systems: Case Study in Smart Microgrids and 5G RAN” won the 2023 Faculty of Engineering’s Best Doctoral Thesis Award at University of Ottawa.
\end{IEEEbiography}

\begin{IEEEbiography}[{\includegraphics[width=1in, height =1.25in,clip,keepaspectratio]{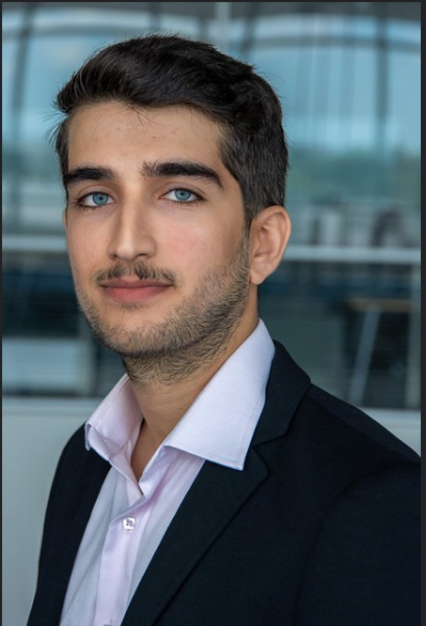}}]{Seyedsina Nabavirazavi}
Seyedsina Nabavirazavi is a Ph.D. candidate in Computer Science at Florida International University. His research explores the security and privacy challenges of machine learning, with a particular focus on federated learning and large language models. His dissertation develops a comprehensive framework to enhance the robustness and resilience of these systems against emerging threats. He holds a B.Sc. in Computer Engineering from the University of Tehran.
\end{IEEEbiography}

\begin{IEEEbiography}[{\includegraphics[width=1in, height =1.25in,clip,keepaspectratio]{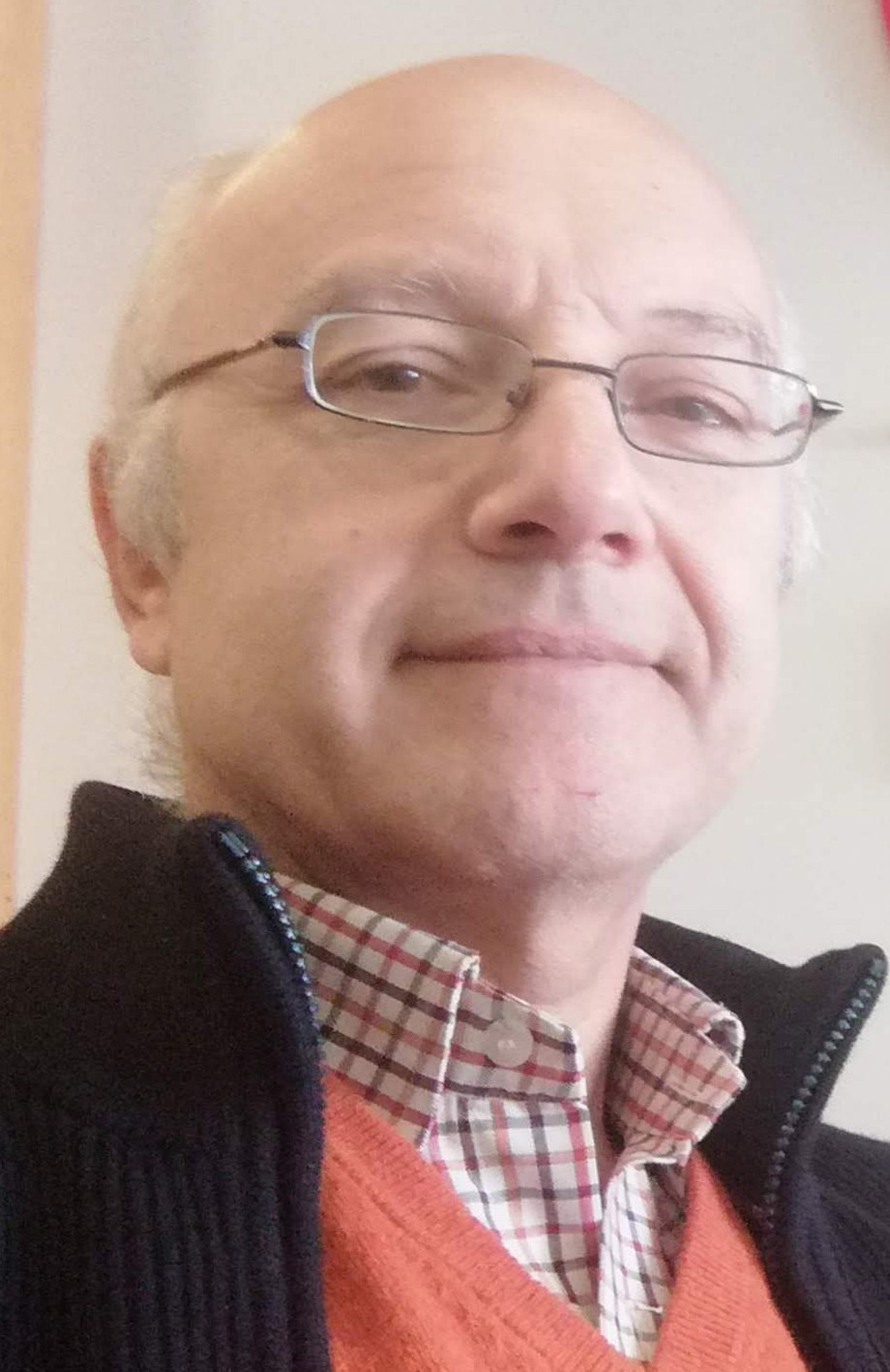}}]{Luis Almeida}
Luis Almeida graduated from the University of Aveiro in Portugal and is currently Full Professor in the Electrical and Computer Engineering Department of the Faculty of Engineering of the University of Porto, Portugal, where he coordinates the Distributed and Real-time Embedded Systems laboratory (DaRTES). He was Vice-Director of the CISTER Research Center on Real-Time and Embedded Computing Systems (2018-2025), Past Chair of TCRTS - the IEEE Technical Committee on Real-Time Systems (Chair in 2020-2021) and Chair of EMSIG - the EDAA Special Interest Group on Embedded Systems. He is Editor-in-Chief of the Springer Journal of Real-Time Systems and Associate Editor of the Elsevier Journal of Systems Architecture. He was Program and General Chair of the IEEE Real-Time Systems Symposium in 2011 and 2012, respectively, and Local co-Chair in 2016, as well as General co-Chair of CPSweek 2018. He was also Trustee of the RoboCup Federation from 2008 to 2016 and Vice-President from 2011 to 2013. He participated in numerous funded national and international research projects and regularly participates in the organization of scientific events in real-time communications for distributed industrial/embedded systems, for teams of cooperating agents and for the Internet-of-Things.
\end{IEEEbiography}
\end{document}